\documentclass[letterpaper]{article} 
\usepackage{aaai2027}  
\nocopyright
\usepackage[hyphens]{url}  
\usepackage{graphicx} 
\usepackage{amsfonts}
\usepackage{amsmath}
\usepackage{tcolorbox}

\usepackage{natbib}  
\usepackage{caption} 
\usepackage{algorithm}
\usepackage{algorithmic}
\usepackage{makecell}
\usepackage{multirow}
\usepackage{newfloat}
\usepackage{xspace}
\usepackage{listings}
\DeclareCaptionStyle{ruled}{labelfont=normalfont,labelsep=colon,strut=off} 
\floatstyle{ruled}
\newfloat{listing}{tb}{lst}{}
\floatname{listing}{Listing}

\usepackage{booktabs}

\title{The Clinician's Veto: Navigating Trust, Liability, and Uncertainty in Autonomous AI Prescribing}

\author{
    Eileanor LaRocco\textsuperscript{\rm 1},
    Sarah Tan\textsuperscript{\rm 2},
    Adarsh Subbaswamy\textsuperscript{\rm 3},
    Anne Andrews\textsuperscript{\rm 4},
    Andrew Taylor\textsuperscript{\rm 1},\\
    \textbf{Cree Gaskin\textsuperscript{\rm 1}},
    \textbf{Chirag Agarwal\textsuperscript{\rm 1}}
}
\affiliations{
    \textsuperscript{\rm 1}University of Virginia\\
    \textsuperscript{\rm 2}Cornell University\\
    \textsuperscript{\rm 3}University of Maryland, Baltimore\\
    \textsuperscript{\rm 4}National Institute of Standards and Technology\\
}

\newcommand{\xhdr}[1]{\noindent{{\bf #1.}}}
\newcommand{\ie}{\textit{i.e., \xspace}}
\newcommand{\eg}{\textit{e.g., \xspace}}

\newcommand{\hide}[1]{}

\newcommand{\revision}[1]{{\color{red}#1}}

\usepackage{colortbl}

\definecolor{Gray}{gray}{0.9}
\definecolor{LightCyan}{rgb}{0.88,1,1}
\definecolor{darkred}{rgb}{0.8,0.1,0.1}
\definecolor{darkyellow}{rgb}{0.95, 0.68, 0.22}
\definecolor{darkgreen}{rgb}{0.1,0.8,0.1}
\newcolumntype{a}{>{\columncolor{Gray}}c}
\newcolumntype{b}{>{\columncolor{white}}c}

\begin{document}

\maketitle
\begin{abstract}
\looseness=-1 Autonomous AI systems are transitioning from advisory roles to autonomous ones for medication prescriptions. Recent U.S. bill H.R. 238 and Utah's prescription-renewal pilot program both authorize AI to prescribe medications in an agentic capacity.  While many regulatory guidelines suggest aggregate model performance metrics at the point of clearance, they do not require i) calibrated per-prediction confidence for action-gated thresholds, ii) differentiated communication between uncertainty arising from model ignorance (epistemic) from genuine clinical ambiguity (aleatoric), and iii) inferential transparency at the moment of decision enabling liability allocation. Here, we argue these three architectural features are minimum conditions for safe autonomous prescribing, and validate them with a survey of \textbf{136} U.S. prescribing clinicians. Our results suggest prescribing clinicians i) would not permit autonomous prescribing without a confidence-based escalation mechanism, ii) preferred a competing-options summary for aleatoric uncertainty but preferred abstention for epistemic uncertainty, and iii) were \textbf{only} willing to accept liability when inferential transparency enabled them to make a decision under acknowledged uncertainty. These findings indicate that our recommended architectural features would encourage higher rates of clinician adoption of autonomous AI prescribing, largely through collapsing much of what ``autonomy'' conventionally means. 
As legislation and pilot programs proceed, our proposed requirements provide opportunities for regulation to constrain the degree of autonomy that should be ethically granted to AI in prescribing while aligning liability with the institutional actors who control system design and deployment.
\end{abstract}


\section{Introduction}

\begin{figure*}[t]
    \centering
    \includegraphics[width=0.81\textwidth]{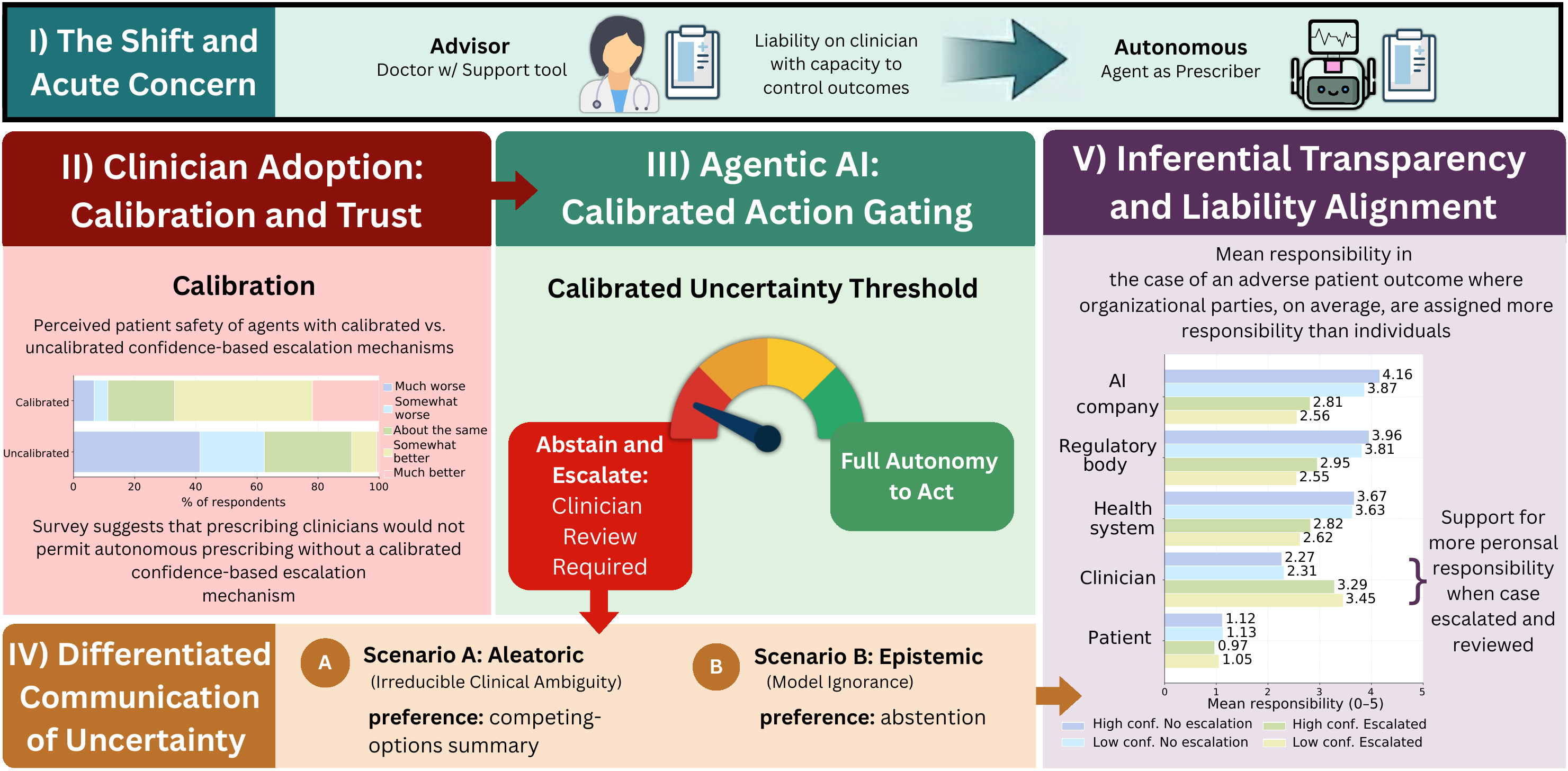}
    \caption{\looseness=-1\xhdr{A Clinician-driven Case for Constraining Autonomous AI in Prescribing} \textbf{I)} As prescribing shifts from Clinicians with AI as support tools to autonomous AI agents, liability risks falling on clinicians who can no longer control outcomes. \textbf{II)} We argue calibration is the precondition that enables safe \textbf{action gating}: a calibrated confidence threshold lets the agent act autonomously when confident but abstain and escalate for clinician review when not. \textbf{III)} Our survey of \textbf{136} U.S. prescribing clinicians show that clinicians rated calibrated escalation as substantially safer than uncalibrated, with a majority unwilling to permit autonomous prescribing without it. \textbf{IV)} Upon escalation, the appropriate content depends on the \textbf{differentiated communication of uncertainty}: clinicians preferred a competing-options summary when uncertainty was aleatoric but shifted toward abstention when it was epistemic. \textbf{V)} Finally, \textbf{inferential transparency} at the point of decision supports \textbf{liability alignment}: across adverse-outcome scenarios, organizational actors were assigned more responsibility than individuals, with clinicians accepting more personal responsibility when an escalated case was surfaced for their review.}
    \label{fig:placeholder}
\end{figure*}

\looseness=-1 The deployment of autonomous artificial intelligence (AI) in healthcare has become a live policy question. At the federal level, H.R. 238 (the Healthy Technology Act of 2025) proposes that AI systems can qualify as legally recognized prescribers \cite{rep_schweikert_text_2025}, while Utah is piloting a program covering 192 drugs in which an autonomous AI agent (software capable of multi-step tasks, decisions, and actions with little to no human supervision) processes prescription renewals for chronic conditions~\cite{mello_utahs_2026}. Together, these developments signal a prioritization of autonomous AI prescribing ahead of the technical and ethical conditions for its safe deployment.

Existing frameworks focus on advisory decision-support environments where a human clinician retains decision-making authority~\cite{grote_ethics_2020, mittelstadt_ethics_2016}. To date, evaluation criterion of fully autonomous prescribing agents are absent from the literature, leaving the field without governance guardrails for agentic action~\cite{damiani_potentiality_2023}. Consequently, prior work has not, to our knowledge, operationalized the precise architectural safeguards required to constrain autonomous prescribing, creating a structural vulnerability where liability is displaced onto clinicians who lack genuine capacity for oversight.

\looseness=-1 To address this gap, we present, to our knowledge, the first clinician-validated framework of minimum architectural requirements for autonomous prescribing. We demonstrate that safe deployment requires: (1) calibrated, action-gated confidence thresholds that enforce abstention on low-confidence decisions; (2) differentiated uncertainty communication that explicitly separates epistemic uncertainty (model ignorance from sparse or biased training data) from aleatoric uncertainty (irreducible clinical ambiguity); and (3) inferential transparency at the point of care that renders individual autonomous actions auditable and enables fair liability allocation. Our survey validates these requirements: clinicians reject uncalibrated autonomous prescribing, differentiate preferred interface content by uncertainty source, and assign more liability to organizations than individuals when AI is fully autonomous (does not escalate to clinician review) but accept more liability when AI uncertainty is communicated.

\looseness=-1 We posit that satisfying these minimal conditions collapses much of what ``autonomy'' conventionally means. A system that abstains on low-confidence cases, gates clinician involvement by uncertainty type, and exposes its reasoning for auditing resembles a heavily supervised decision-support tool more than an autonomous agent. This collapse is \textbf{central} to our argument: \textit{the minimum safety conditions in public health are stringent enough to constrain how much autonomy can ethically be granted}. As U.S. policy is already advancing toward autonomous prescribing despite unresolved technical and ethical risks, establishing the \textbf{minimum architectural safeguards} required for safe deployment provides an immediate, actionable framework to protect patient safety and realign liability.
Although we study prescribing, the structure is general: an agent used for high-stakes tasks, gated by its own uncertainty estimates, embedded in institutions that must allocate liability for its errors -- a configuration that applies wherever agentic AI is delegated real authority.

\section{Related Work}
Our work lies at the intersection of public health regulation, clinician trust, and autonomous AI governance.

\looseness=-1\xhdr{Policy Landscape} The Food and Drug Administration (FDA) in U.S. approves drugs and medical devices (including software performing a medical function) to assure their safety, but it does not regulate the practice of medicine \cite{fda_what_2026, fda_device_2024}. In January 2025, the FDA released draft guidance for AI-Enabled Device Software Functions acknowledging that ``\textit{AI-enabled devices span a continuum of decision-making roles}'', but stopping short of addressing fully autonomous systems by stating that they ``\textit{rely on the human to interpret the AI outputs and ultimately make clinical decisions}''~\cite{fda_artificial_2025}. In January 2026, the FDA released new guidance for clinical decision support (CDS) software, drawing a clearer line between non-device and device CDS, with autonomous agents software falling under FDA regulation~\cite{schellhous_fda_2026}. However, \textbf{no autonomous AI prescription services have been cleared} by the FDA to date~\cite{teng_autonomous_2026}. H.R. 238 is the first legislation to address autonomous AI prescribing, proposing that AI ``\textit{can qualify as a practitioner eligible to prescribe drugs}'' if authorized by the State or FDA. Shortly afterward, Utah announced their pilot program with Doctronic, raising clinician questions about algorithmic recommendations, liability, predicate creep, and patient needs~\cite{mello_utahs_2026}. While \citet{gilbert_consternation_2025} note that autonomous prescribing systems are not inherently problematic provided they are effective, safe, and serve patient interests, unresolved accountability and liability questions \textbf{remain unaddressed}.\vspace{0.05in}


\looseness=-1\xhdr{Evidence Base on Autonomous AI Prescribing} A 2023 systematic review of AI in primary care medication found no studies at higher autonomy levels, with existing systems limited to assistive roles; 71\% reported reductions in preventable errors, including overprescribing and adverse drug interactions \cite{damiani_potentiality_2023}. The only published prospective trial of autonomous AI prescribing of which we are aware is daGOAT \cite{chen_autonomous_2025}, where an algorithm analyzed lab data and, subject to physician veto, prescribed an immunosuppressant for graft-vs.-host disease. 

\looseness=-1\xhdr{Ethical Stakes of Autonomous Prescribing} The ethical stakes of autonomous prescribing differ from advisory prescribing. Work on algorithmic ethics identifies epistemic, normative, and traceability concerns \cite{mittelstadt_ethics_2016, char_implementing_2018, morley_ethics_2020}, each intensifying as systems move from supporting clinician judgment to substituting for it.~\citet{grote_ethics_2020} argue that medical AI entails epistemic and normative trade-offs that challenge accountability norms. In advisory settings, the clinician's decision-making authority preserves accountability, but \textbf{autonomy removes that mediation by design}.

\looseness=-1 H.R. 238 and the Utah pilot allow AI to prescribe without clinician oversight. Yet our own and prior surveys suggest autonomous AI is unlikely to operate without at least nominal clinician sign-off \cite{american_medical_association_physician_2026, busch_multinational_2025}. This yields in which the AI acts and the clinician bears liability for adverse outcomes. This is the predictable consequence of building autonomous systems into oversight structures designed for advisory ones, raising the justice-and-accountability concern consistently flagged when responsibility is distributed across actors with asymmetric control \cite{mittelstadt_ethics_2016}.

\section{Safe Autonomous AI Prescribing}

\looseness=-1 We test three hypotheses about these requirements with our survey of 136 U.S. prescribing clinicians.

\subsection{Calibrated, Action-Gated Confidence Thresholds}
\looseness=-1 The first requirement is calibrated confidence thresholds that act as a safety gate: an autonomous AI agent must abstain from making a decision when its confidence is low. This is critical because, unlike advisory systems, the clinician is structurally absent from the initial decision, making the model's confidence estimate the sole mechanism for deciding whether to escalate to clinician review.

\noindent\textbf{Hypothesis 1:} Clinicians will treat a calibrated confidence-based escalation mechanism as necessary for safe autonomous AI prescribing. Specifically, i) a majority will not support autonomous prescribing without such a mechanism, and ii) clinicians will rate calibrated systems as meaningfully safer than uncalibrated ones.

\looseness=-1 We treat uncertainty and confidence as inverse, \ie high uncertainty = low confidence. Calibrated probabilities represent real-world uncertainty estimates where higher probabilities = higher confidence \cite{lambert_trustworthy_2024}. Neural networks are systematically miscalibrated~\cite{guo_calibration_2017,van2019calibration}, and LLMs in particular are overconfident~\cite{tanneru2024quantifying}: their probabilities are typically unreliable and miscalibrated, constituting a safety hazard \cite{bengio_superintelligent_2025}. We therefore \textbf{argue for transparent alignment of confidence with accuracy} through communication of calibrated uncertainty. \citet{bhatt_uncertainty_2021} established calibration as the precondition for using uncertainty as a transparency measure. In autonomous prescribing, confidence becomes load-bearing safety infrastructure serving three functions absent from advisory systems: action gating, real-time detection of epistemic uncertainty, and preserving an auditable trail for liability allocation.


\looseness=-1 
\citet{kompa_second_2021} found that four widely cited deployed medical ML models lack any abstention mechanism. A model that \textbf{cannot abstain cannot gate}: a non-abstaining system with a mis-calibrated gate will prescribe even when escalation is warranted, whereas a well-calibrated model uncertain about a correct one appropriately defers. The agentic architecture proposed in \citet{srinivasu_exploring_2026} makes this concrete: their framework includes an Escalation Unit that suspends action when confidence falls below a threshold.

\subsection{Differentiated Uncertainty Communication}

\looseness=-1 The second requirement is that the system must differentiate its uncertainty source when escalating. The content displayed to the clinician \textbf{should reflect the distinction between epistemic and aleatoric uncertainty}, as the appropriate clinician response differs based on whether the model is limited or the case is genuinely ambiguous.

\looseness=-1\noindent\textbf{Hypothesis 2:} When an AI agent escalates a case, clinicians will determine that the output content should differ by source of uncertainty, \ie a competing-options summary is appropriate under high aleatoric uncertainty but no AI recommendation should be shown under high epistemic uncertainty. 

\textbf{Aleatoric uncertainty} reflects irreducible variability inherent to the data-generating process—a genuinely ambiguous clinical presentation for which multiple evidence-based treatment options exist and the literature does not clearly favor one. \textbf{Epistemic uncertainty} reflects the system's uncertainty not when the case is inherently hard, but when it has not seen enough examples like it \cite{hora_aleatory_1996, bhatt_uncertainty_2021}. The technical case for separating these uncertainties is established, it has not been examined whether clinicians act on the distinction when receiving an escalated case \cite{fan_position_2025, liu_role_2025}. Once a case is escalated and reaches the clinician as decision support, the source of uncertainty becomes directly relevant to what the interface should display. In aleatoric cases, a competing-options summary is appropriate: grounded in real data, it preserves analytical value while deferring final judgment to the clinician. In epistemic cases, autonomous action should halt entirely and \textbf{no AI recommendation} should be displayed, since any output falls outside the validated distribution. This guards against automation bias: clinicians' tendency to defer to an AI's recommendation even when it is incorrect \cite{kucking_automation_2024, abbott_understanding_2025, dratsch_automation_2023, goddard_automation_2012, wickens_complacency_2015, khera_automation_2023, qazi_automation_2026}. Under epistemic uncertainty, the model has no validated basis for any recommendation, and displaying a preferred option is precisely when automation bias is most damaging \cite{liu_role_2025}. Suppressing the recommendation functions as a structural guard rather than a mere communication preference.\vspace{0.05in}

\subsection{Inferential Transparency and Liability Allocation}

The final requirement is inferential transparency; a traceable reasoning record necessary for real-time audit, action gating, and liability allocation after an adverse outcome.

\noindent\textbf{Hypothesis 3:} Following an adverse outcome from an autonomous AI prescribing decision, clinicians will assign more liability to higher-level oversight organizations (health systems, regulatory bodies, AI companies) than to individuals (physician, patient). This organizational-vs-individual gap will narrow when the agent communicated its confidence through inferential transparency.

\looseness=-1 Here, we differentiate two terms, arguing that a fundamental regulatory gap exists between them: \textbf{1) Documentation transparency:} aggregate model performance communicated at deployment; and \textbf{2) Inferential transparency:} per-prediction transparency communicated at each clinical decision. In advisory AI, explainability and uncertainty communication primarily support clinician decision \cite{alelyani_validated_2025}. In autonomous AI, the question becomes whether explanations are faithful, traceable, and exposed through interfaces in a way that supports audit. Alelyani et al. note that generative AI introduces auditability challenges 
that require explainability and monitoring to be treated as dynamic rather than static. Under NIST's AI Risk Management Framework, transparency addresses what happened, explainability addresses how a decision was made, and interpretability addresses why \cite{tabassi_artificial_2023}. Current regulations do not close the auditability gap as they are designed almost exclusively around documentation transparency. The FDA's 2025 draft guidance relies on model cards, subgroup analyses, and aggregate accuracy which, while better than opaque deployment \cite{gilbert_could_2025}, \textbf{do not capture} what matters most for safe individual decisions. Agentic systems worsen this gap since model cards are static artifacts produced before deployment while agents adapt to new populations over time.~\citet{gardhouse_regulating_2026} note that many agentic risks only surface in specific deployment contexts and keep evolving. Requiring documentation transparency at release while omitting inferential transparency during operation means the \textbf{regulatory record reflects the model as validated, not as it is currently behaving.}

\looseness=-1 For liability, the ``many-hands problem'' \citet{gardhouse_regulating_2026} \textbf{risks positioning clinicians as the de facto accountability layer for autonomous AI actions} they cannot meaningfully verify. Outcomes following aleatoric uncertainty more naturally rest with the clinician who made the final call, while outcomes following epistemic uncertainty more plausibly rest with the developer whose calibration failed to bound the system's competence. Regulation treating all uncertainty equivalently cannot assign liability in ways that create appropriate incentives. Numerous technical approaches for separating these uncertainty types are available, including proper scoring rules \cite{hofman_quantifying_2024}, direct quantification in neural networks \cite{jones_direct_2022}, and hyper-diffusion models \cite{chan_estimating_2024}.

This distinction is equally crucial for auditing agents in the wild. Epistemic uncertainty should decrease as training data grows while aleatoric uncertainty should not, so a system tracking only aggregate uncertainty cannot distinguish model degradation from irreducible clinical complexity. Both layers are necessary; documentation transparency to establish validation before deployment, and inferential transparency to ensure each autonomous action can be examined, challenged, and attributed. Mandating only the former creates the appearance of \textbf{oversight without substance.}

\section{Survey Methods}

\looseness=-1 Our IRB-approved survey asked prescribing clinicians for their opinions on autonomous AI agents as prescribers (\ie AI that can issue prescriptions without a physician reviewing each individual decision). Participants completed a short online survey administered via Qualtrics and disseminated through multi-institutional email lists. No personally identifiable information was collected. To qualify, participants must be 18 or older, English-speaking, and currently hold U.S. prescribing authority, including but not limited to physicians, physician assistants, and nurse practitioners. Refer to Appendix C for details on full test specifications, multiplicity handling, and missing data.

\looseness=-1 We organized the survey into six blocks: \textbf{demographics} (Q1–Q4), \textbf{baseline attitudes} (Q5–Q9), H1: \textbf{calibrated, action-gated confidence thresholds} (Q10–Q12), H2: \textbf{differentiated uncertainty communication} (Q13–Q15), H3: \textbf{inferential transparency and liability allocation} (Q16–Q17), and a final optional open-ended question (Q18). Demographic items used single-select multiple choice (role, specialty, prescriptive authority, years prescribing), AI familiarity was captured on five-point Likert-type scales (Q5–Q6), and a multi-select checklist recorded the tasks for which respondents use AI (Q7). Baseline support for autonomous prescribing was measured with two parallel three-point (oppose/neutral/support) matrix items spanning four drug-risk categories, for non-complex (Q8) and medically complex (Q9) patients. Hypothesis 1 was tested with a four-point endorsement of calibrated escalation (Q10) and a five-point Likert-type safety-rating matrix (much worse$\to$much better) comparing a calibrated and an uncalibrated agent against a no-escalation baseline (Q11). Hypothesis 2 presented matched aleatoric and epistemic uncertainty vignettes followed by single-select items on whether the scenarios differ (Q13) and what the escalated interface should display in each (Q14–Q15). Hypothesis 3 used four adverse-outcome vignettes crossing AI confidence (low/high) with escalation behavior (no escalation/communication vs. escalation with clinician concurrence), each asking respondents to rate the responsibility of five parties on a 0–5 scale (Q16A–Q17B). Free-text ``Other (please specify)'' options accompanied most categorical items and were recoded into existing categories if they mapped cleanly or recoded as new categories. The final optional open-ended question (Q18) invited clinicians to identify issues the AI community is not adequately addressing, capturing qualitative feedback beyond the structured items. Please refer to Appendix A for the full survey instrument.

\section{Results}

\looseness=-1 We first summarize the sample, then report results for H1–H3 and the qualitative analysis.
Please refer to Appendix C for additional analyses on subpopulation break-outs.\vspace{0.05in}

\xhdr{Descriptive Analysis} Overall, physicians comprised 75\% of our respondents, with the remainder being nurse practitioners (22.1\%) and physician assistants (2.9\%). Internal medicine represented 26.5\% of our sample, with surgery, pediatrics, and family medicine/general practice representing 12.5\%, 11.0\%, and 9.6\%, respectively. Most respondents held full independent prescriptive authority (77.2\%) with the distribution for any level of prescribing privileges as: 1-5 years (30.1\%), 6-10 years (16.2\%), 11-15 years (17.6\%), 16-20 years (10.3\%), and $\operatorname{>}$ 20 years (25.7\%). A consolidated summary of the sample's demographic composition (role, specialty, prescriptive authority, years prescribing, and current AI use) is provided in Appendix B. Familiarity with AI was overall high: Generative AI (\textbf{96.3\%}) and Agentic AI (83.1\%), and \textbf{86\%} used AI in their current role. See Fig.~\ref{fig:1} in Appendix B for further details.\vspace{0.05in}

\begin{figure}[h]
    \centering
    \includegraphics[width=0.9\columnwidth]{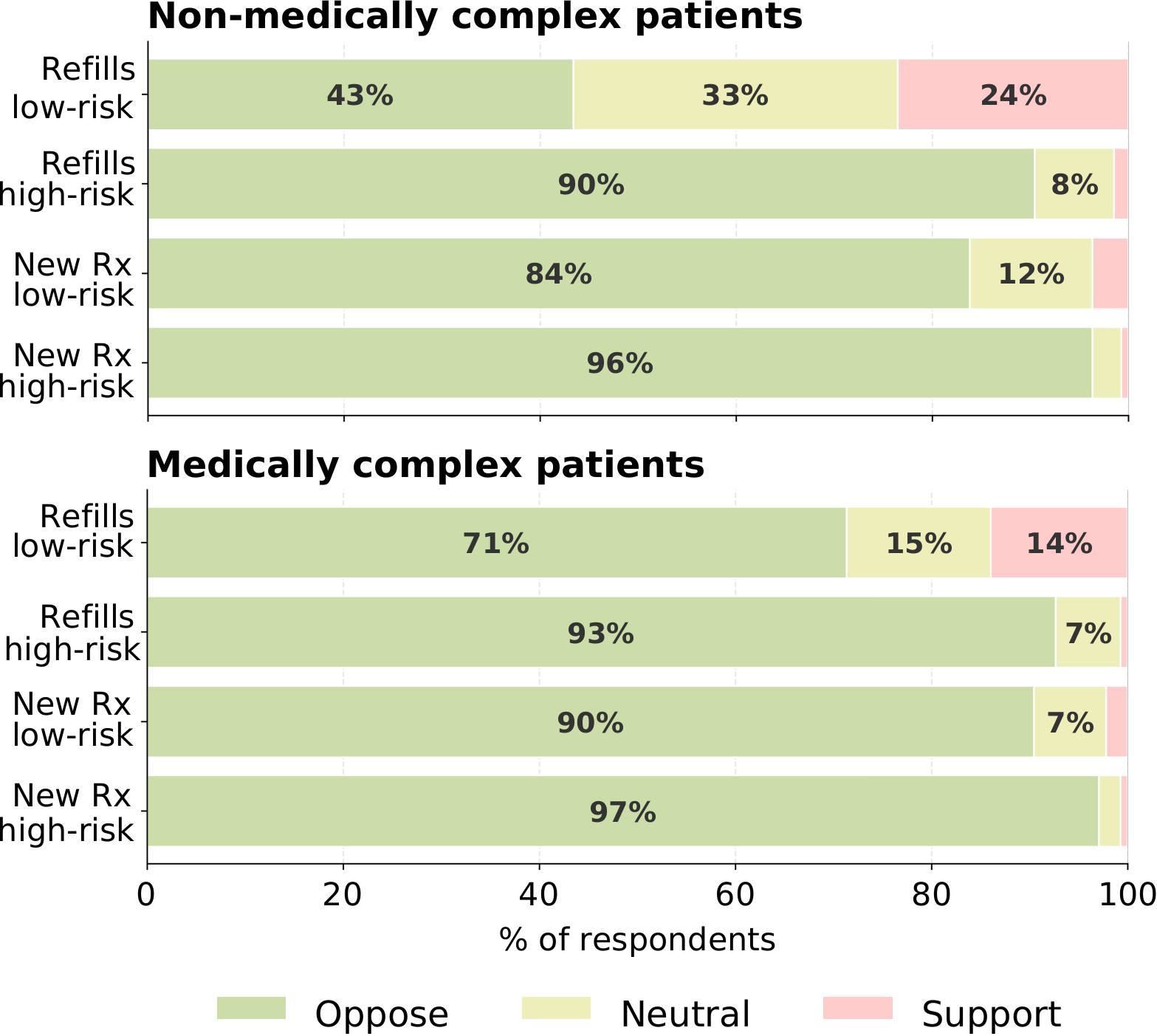}
    \vspace{-0.05in}
    \caption{Clinician support for autonomous AI prescribing without requiring clinician sign-off by drug category and patient complexity (non-medically complex patients (\textbf{top}), medically complex patients (\textbf{bottom}). Results show that opposition exceeds \textbf{83\%} for all new prescriptions and high-risk refills across both patient types, with the only non-trivial support shown for low-risk refills.}\vspace{-0.05in}
    \label{fig:2}
\end{figure}

\noindent\looseness=-1\textbf{Do clinicians support Autonomous AI prescribing?}
On average, we found that the s\textbf{upport for autonomous AI prescribing was very low} across all drug categories. For new prescriptions of both low and high-risk drugs and refills of high-risk drugs, support was at/under 5\% for both medically complex patients and non-complex patients. The only category showing meaningful (though still minor) support was refills of low-risk drugs for non-complex patients (23.5\% vs. 14\% for complex patients; Fig.~\ref{fig:2}).

\subsection{Calibrated, Action-Gated Confidence Thresholds}

\looseness=-1 \textbf{Setup.} Clinicians were given the definition of a calibrated confidence-based escalation mechanism (\ie a mechanism where the agent automatically halts and routes a case to clinician review when its confidence falls below a defined threshold; see Appendix~A Q10 for more details), and asked 1) if the presence of said mechanism affected their willingness to permit AI agents to prescribe drugs (H1(a)); 2) to compare the patient safety of two AI agents --- one with a calibrated confidence-based escalation mechanism, and one without --- relative to an AI agent with no escalation mechanism at all (H1(b)). They were then asked who should bear primary responsibility for setting the threshold for the mechanism.

\begin{figure}[h]
    \centering
    \includegraphics[width=0.9\columnwidth]{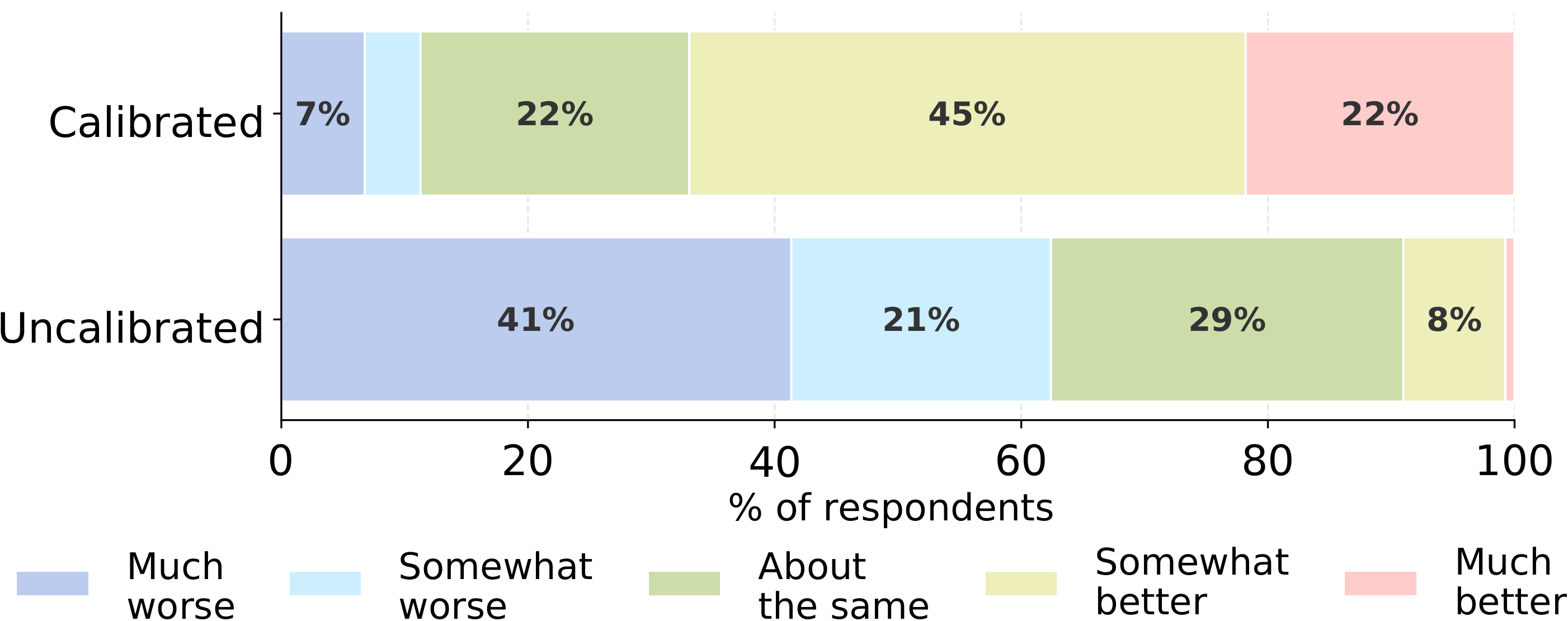}\vspace{-0.05in}
    \caption{\looseness=-1 Rated patient safety of agents with calibrated vs. uncalibrated confidence-based escalation mechanisms, each relative to an agent with no escalation. Clinicians rated the calibrated agent as \textbf{safer} than the uncalibrated agent.}
    \label{fig:3}
\end{figure}

\noindent\textbf{Findings.} (H1(a)) For the effect of the mechanism on their willingness to permit AI agents to prescribe drugs, 67.4\% of clinicians chose the \textbf{strongest endorsement} (\textit{``To a great extent — I would not permit AI agents to prescribe drugs without it''}). This was our anchored primary test for H1(a) (one-sided binomial vs.\ $0.5$, p $\operatorname{<}$ 0.001). With a supporting descriptive cut of the same item, 84.4\% chose one of the top two response options (p $\operatorname{<}$ 0.001). (H1(b)) For the two agent scenario, clinicians rated the patient safety of the uncalibrated agent (mean 2.06, SD 1.05) to be significantly worse than the calibrated agent (mean 3.71, SD 1.07; p $\operatorname{<}$ 0.001) relative to an agent with no escalation mechanism at all (Figure~\ref{fig:3}).

\looseness=-1 Further, there was no clear consensus from the clinicians on who should set the confidence threshold at which an AI agent escalates to clinician review. When asked who should bear primary responsibility for setting the confidence threshold, no single party was given majority support, however the \textbf{organization-level categories combined accounted for 59.5\%} (regulatory body, regulatory body with clinician review/adjustment, independent clinical board, and multi-stakeholder body). Only \textbf{5.3\%} endorsed AI-companies to do so. See Figure~\ref{fig:4} in Appendix B for additional details.

\subsection{Differentiated Uncertainty Communication}

\looseness=-1\textbf{Setup.} Although clinicians were not given the names of the types of uncertainty, clinicians were presented with scenarios representing aleatoric uncertainty (Scenario A) and epistemic uncertainty (Scenario B).

\begin{figure}[t]
    \centering
    \includegraphics[width=0.9\columnwidth]{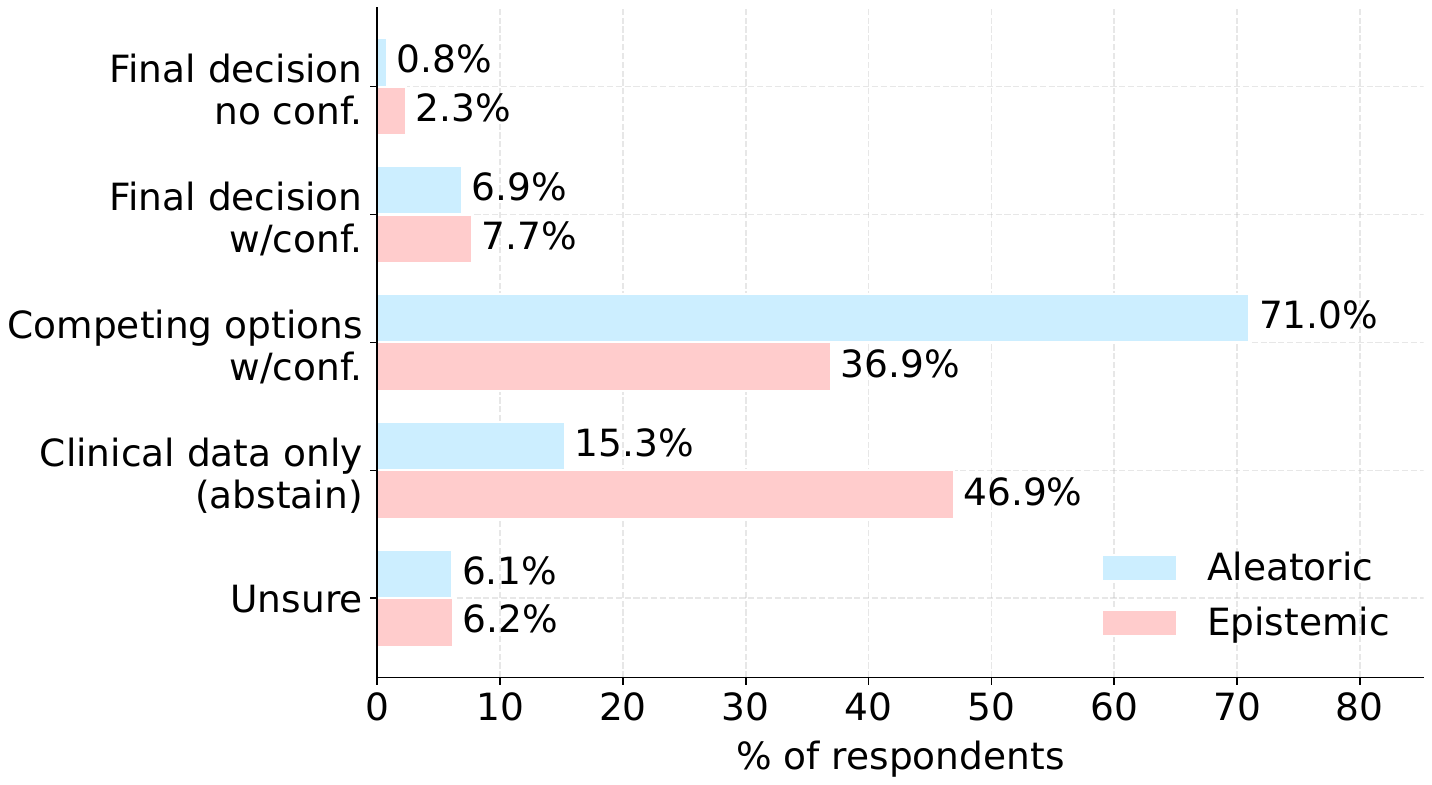}\vspace{-0.075in}
    \caption{Preferred content of the escalated interface by source of uncertainty (aleatoric vs. epistemic). Results show that clinicians preferred to be shown competing options with associated confidence scores when aleatoric uncertainty was high, but when epistemic uncertainty was high, the preference shifted towards wanting the agent to abstain from recommending and present the clinical data only.}
    \label{fig:5}
\end{figure}

\looseness=-1\noindent\textbf{Findings.} Clinicians responded primarily that the two scenarios require fundamentally different approaches to review the case (59.8\%) with an additional 9.8\% viewing the scenarios as having only minor differences; 22.7\% would approach both the same way and 7.6\% were unsure. A majority (69.6\%) thus perceived at least some difference between the aleatoric and epistemic presentations. Preferences for the content of the escalated interface aligned mostly with H2. While 71.0\% preferred a competing-options summary and 15.3\% preferred abstention (Fig.~\ref{fig:5}) in aleatoric scenario, the \textbf{distribution shifted significantly}: abstention rose to 46.9\% and competing-options dropped to 36.9\% (p $\operatorname{<}$ 0.001 for both) for the epistemic scenario. Because H2 is assessed with two McNemar contrasts (the competing-options shift and the abstention shift), we treat them as a single family and apply Holm--Bonferroni correction; both remain significant after correction (Holm-adjusted p $\operatorname{<}$ 0.001). While H2 holds, the epistemic scenario does not produce a majority endorsement of abstention, but instead shows split preference between abstention and competing-options, with some free-response indicating preference for both.

\subsection{Inferential Transparency and Liability Allocation}

\looseness=-1\textbf{Setup.} We asked clinicians to allocate liability across five parties (organization who developed the AI agent, regulatory body who cleared the agent, health system that deployed the agent, the patient's clinician, and the patient who consented to use of the agent) given a scenario where an agent issued a new prescription and the patient experienced an adverse drug reaction. This was tested across four scenarios with varying agent confidence (low/high) and escalation behavior (escalation with clinician vs. no escalation to the clinician).\vspace{0.05in}

\looseness=-1 \noindent\textbf{Scenarios 1 \& 2.} An AI agent issued a new prescription. The AI agent had {low- (Scenario 1) or high-confidence (Scenario 2)} in its decision, \textcolor{red!50!black!75}{\textbf{did not communicate}} its confidence level to the clinician, and \textcolor{red!50!black!75}{\textbf{did not trigger escalation}} for review. The patient experienced a serious adverse drug reaction.\vspace{0.05in}

\looseness=-1\noindent\textbf{Scenarios 3 \& 4.} An AI agent issued a new prescription. The AI agent had {low- (Scenario 3) or high-confidence (Scenario 4)} in its decision, \textcolor{green!50!black!100}{\textbf{did communicate}} the confidence level to the clinician, and \textcolor{green!50!black!100}{\textbf{did trigger escalation}} for review. The \textcolor{green!50!black!100}{\textbf{clinician concurred and documented why}}. The patient experienced a serious adverse drug reaction.\vspace{0.05in}


\looseness=-1\noindent\textbf{Findings.} Consistent with H3, mean responsibility assigned to the organization was significantly higher than that assigned to the individuals in all four scenarios above. In scenarios 1, 2, and 4, the difference was significant (p $\operatorname{<}$ 0.001). In Scenario 3, the difference was still significant (p = 0.017), but not as high, reflecting the ambiguity of the situation and a partial shift of responsibility toward the overseeing clinician. Because the predetermined org-vs.-individual contrast is evaluated once per scenario, the four tests form a single family and are corrected together with Holm--Bonferroni; all four remain significant at the $0.05$ level after correction (the three strongest at Holm-adjusted p $\operatorname{<}$ 0.005, and Scenario~3, as the largest raw $p$ in the family, essentially unchanged at Holm-adjusted p = 0.017).

\begin{figure}[t]
    \centering
    \includegraphics[width=0.75\columnwidth]{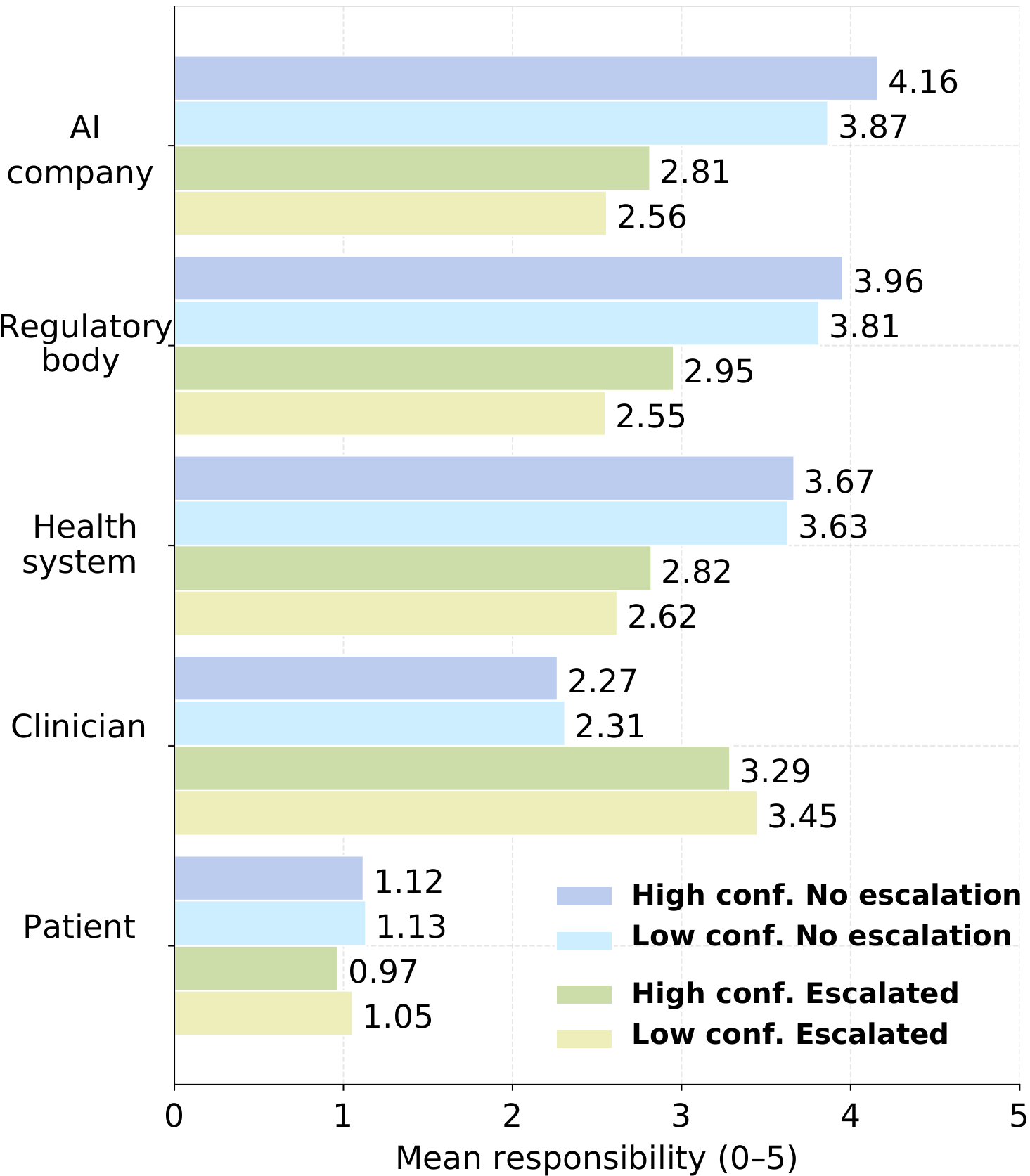}
    \vspace{-0.1in}
    \caption{\looseness=-1 Bar graph of mean responsibility (from 0-5) assigned to each party across four scenarios in the case of an adverse outcome. Results show that organizational parties are assigned more responsibility than individuals in every scenario except when the agent escalates, with patients being assigned the least responsibility throughout despite consent.}
    \label{fig:6}
\end{figure}

More exploratory comparisons revealed two patterns of interest. In the no-escalation scenarios (1 \& 2), AI organization were assigned more responsibility when the model had been highly confident than when it had been low-confident ($\Delta$mean = + 0.29, p = 0.004) — consistent with treating high-confidence-without-escalation when the model was incorrect as a severe architectural failure than when the model accurately assessed its low confidence. In the escalation-and-concurrence scenarios (3 \& 4), overseeing clinicians were assigned less responsibility when the AI had been highly confident than when it had been low-confident ($\Delta$mean = - 0.16, p $\operatorname{<}$ 0.05), suggesting that \textbf{clinicians accept additional responsibility when inferential transparency enabled them to make a substantive judgment} under acknowledged uncertainty (Fig.~\ref{fig:6}). Furthermore, patients were consistently assigned the lowest responsibility of any party across all four scenarios (Fig.~\ref{fig:6}), even though respondents were explicitly told the patient had consented to AI agent usage. \textbf{Clinicians thus did not treat documented consent as transferring meaningful accountability to the patient}. This combined with the patterns found supporting H3 show that respondents prefer that the institutional actors bear the most responsibility for adverse outcomes also should bear the authority over the architectural decisions that produce them.

\subsection{Qualitative Analysis}

\looseness=-1 For Q18, we asked clinicians for open-ended thoughts on what the AI community is not paying enough attention to with regards to AI agents in prescribing. Out of the 67 received responses, sentiments were roughly balanced: positive (39.1\%), negative (37.5\%), and neutral (23.4\%). 
Overall, we received a diverse set of responses that highlight five central areas of concern regarding autonomous AI in prescribing.\vspace{0.05in}

\noindent\looseness=-1 \textbf{Liability.} The overriding concern was that the liability for AI mistakes (H3) will unfairly default to the clinician, outweighing any potential efficiency gain. Respondents specifically noted that they are unwilling to use AI if they bear full responsibility, suggesting compensation is necessary if liability is transferred to them — \textit{``AI, like humans, will make mistakes. Who gets blamed? If the clinician is held responsible at all, then I would never want to use AI. It's just another source of liability with limited improvement in patient care;'' ``If providers are taking liability, they should be compensated.''}\vspace{0.05in}

\noindent\looseness=-1 \textbf{Lack of Ability.} Here, the clinicians raised concerns around the lack of ability of current models. In particular, they doubted the ability of AI agents to match clinical judgment, citing known issues like hallucinations, poor handling of dosing and drug interactions, and an inability to account for nuances (\eg body language) or incomplete documentation in patient records — \textit{``Lack of ability of LLMs (can't compare to clinical judgment);'' ``Black boxes with no visible rationale should not be prescribing meds;'' ``medical records are often incomplete or contain information that is not accurate/current. I would have low confidence using just that information to make autonomous medical decisions.''}\vspace{0.05in}

\noindent\looseness=-1 \textbf{Patient Safety.} Clinicians raised concerns around patient safety with the use of AI in prescribing, including patient risk, bias, and the potential for misuse by ``bad actors.'' Clinicians also pointed out that AI adoption might degrade necessary clinical workflows, such as losing track of patient follow-ups and reducing the quality of personalized patient care (\eg shared decision-making). Many felt the potential benefits were not worth the risks — \textit{``If a provider isn't automatically getting the refill requests to sign off on, then they aren't able to track how frequently a patient is being seen/when they are due for follow up;'' ``Benefits do not outweigh risks (Not worth the small potential benefit for patient care).''}\vspace{0.05in}

\noindent\looseness=-1 \textbf{Administrative and Regulatory Scope.} A common suggestion was to confine AI's use to administrative tasks, viewing autonomous prescribing as a lower priority. This is paired with demands for more external safeguards and regulation to govern deployment — \textit{``Want AI to help with admin-only items'' ``More safeguards and regulation needed;'' ``This just seems like a misguided use of resources and not the place we need to try to save clinician time.''}\vspace{0.05in}

\noindent\looseness=-1 \textbf{Lack of Trust.} Many clinicians raised trust issues driven by perceived haste (``too fast/soon'') and concerns over conflicts of interest. Specifically, clinicians fear corporate, pharmaceutical, or political influence shaping the AI platform to push specific interventions (\eg naturopathic treatments). Additionally, a significant number of respondents expressed explicit, unequivocal opposition to AI agents role in the field.  — \textit{``AI agents have NO ROLE in medicine;'' ``there need to be significant safeguards when using AI and I don't trust that it is happening;'' ``would not want MAHA naturopathic interventions to be pushed by the AI platform owner/programmers.''}

\section{Discussion}

\looseness=-1 A further concern motivates our work: \textit{the institutional configuration under which autonomous prescribing infrastructure will be built and operated.} The most prominent cautionary case in commercial health AI to date is the Google DeepMind–Royal Free London NHS arrangement, in which patient data was transferred to a commercial AI developer under terms later found to be inadequate in their legal and ethical basis~\cite{powles_google_2017}. \textbf{Who builds, owns, and profits from autonomous prescribing systems} is not separable from the question of whether those systems can be safely operated (\textcolor{blue!50!black!75}{\textbf{Anonymous respondent: \textit{``something I would be worried about is whether pharma or AI companies are able to influence the AI recommendations in ways that are commercially biased (\ie conflicts of interest)''}}}). If FDA clearance becomes the gateway to prescribing authority, the developers of cleared systems become effective prescribers at scale with commercial entanglements with the pharmaceutical industry. These concerns intensify as algorithmic systems move from supporting clinician judgment to replacing it.\vspace{0.05in}

\noindent\looseness=-1 \textbf{Limitations.} While clinicians reported willingness to accept more liability when working alongside AI with review ability, we did not test whether this would hold if they were given less time to familiarize themselves with the case than they would like (\ie a new patient or a more medically-complex patient). Our respondents, though spanning many specialties, came mostly from a single university-associated hospital system; future studies could draw from a wider range of locations and clinic types. Our inferential-transparency hypothesis was also tested implicitly, via a contrast of liability questions rather than a direct preference question—future work should ask clinicians directly about per-prediction transparency requirements. Lastly, given the topic's polarizing nature and general survey response bias, our results likely carry some negative-direction bias but are not necessarily unfounded given our population's high AI-awareness.

\section{Conclusion}

\looseness=-1 Autonomous AI prescribing is moving ahead without the technical and regulatory infrastructure to use it safely. Here, we argue that calibrated, action-gated confidence thresholds, communication of epistemic vs. aleatoric uncertainty, and inferential transparency are minimum requirements for safe autonomous prescribing that constrain how much autonomy can be ethically granted. This matters for policy as H.R. 238's framing of AI as a ``\textit{practitioner eligible to prescribe}'' grants complete autonomy that neither our requirements nor our clinicians support. Our survey of 136 U.S. prescribing clinicians shows that they treat these constraints as preconditions for working alongside autonomous systems. A majority would not permit autonomous prescribing without a calibrated escalation mechanism; most varied the escalated-interface response by uncertainty source, preferring a competing-options summary under aleatoric uncertainty and shifting significantly toward abstention under epistemic uncertainty; and across all four adverse-outcome scenarios assigned greater responsibility to AI organizations than individuals, accepting more personal responsibility only when the agent communicated confidence and they could review the case. These features prove necessary but not sufficient for clinician adoption: baseline support was low across nearly every drug category, and even with calibration \textbf{clinicians did not broadly endorse autonomous prescribing}.



\newpage
\bibliography{aaai2027}

\clearpage
\newpage
\appendix
\section{Appendix}
\label{app:appendix}

\subsection{Appendix A: Survey Instrument}
\label{app:A}

\subsection*{Consent}

\textbf{Q0.} Please read this study information sheet carefully before you decide to participate in the survey: Consent Form. Do you consent to participate in this survey?

\begin{itemize}
    \item[$\circ$] I consent
    \item[$\circ$] I do not consent
\end{itemize}

\textit{Skip to end of survey if participant does not consent.}

\subsection*{Demographics}

\textbf{Q1.} What option best describes your current role?

\begin{itemize}
    \item[$\circ$] Physician (Non-Psychiatrist)
    \item[$\circ$] Physician (Psychiatrist)
    \item[$\circ$] Physician Assistant
    \item[$\circ$] Nurse Practitioner
    \item[$\circ$] Psychiatric Nurse Practitioner
    \item[$\circ$] Other (please specify):
\end{itemize}

\noindent\textbf{Q2.} What option best describes your specialty?

\begin{itemize}
    \item[$\circ$] Internal Medicine
    \item[$\circ$] Family Medicine/General Practice
    \item[$\circ$] Pediatrics
    \item[$\circ$] Surgery
    \item[$\circ$] Obstetrics and Gynecology
    \item[$\circ$] Psychiatry
    \item[$\circ$] Emergency Medicine
    \item[$\circ$] Anesthesiology
    \item[$\circ$] Radiology
    \item[$\circ$] Podiatry
    \item[$\circ$] Not Applicable
    \item[$\circ$] Other (please specify):
\end{itemize}

\noindent\textbf{Q3.} What level of prescriptive authority do you hold?

\begin{itemize}
    \item[$\circ$] Full independent
    \item[$\circ$] Collaborative, restricted, or limited
    \item[$\circ$] Other (please specify):
\end{itemize}

\noindent\textbf{Q4.} How long have you had prescribing privileges?

\begin{itemize}
    \item[$\circ$] 1--5 years
    \item[$\circ$] 6--10 years
    \item[$\circ$] 11--15 years
    \item[$\circ$] 16--20 years
    \item[$\circ$] $>$20 years
\end{itemize}

\subsection*{Baseline Attitudes}

\textbf{Q5.} How familiar are you with generative AI (AI that produces content such as text, images, videos, audio, and software code in response to user prompts)?

\begin{itemize}
    \item[$\circ$] Not familiar at all (never used)
    \item[$\circ$] Slightly familiar
    \item[$\circ$] Moderately familiar
    \item[$\circ$] Very familiar
    \item[$\circ$] Extremely familiar (use daily)
\end{itemize}

\noindent\textbf{Q6.} How familiar are you with agentic AI (autonomous systems that use large language models to reason, plan, and take multi-step actions to achieve goals with minimal human supervision)?

\begin{itemize}
    \item[$\circ$] Not familiar at all (never used)
    \item[$\circ$] Slightly familiar
    \item[$\circ$] Moderately familiar
    \item[$\circ$] Very familiar
    \item[$\circ$] Extremely familiar (use daily)
\end{itemize}

\noindent\textbf{Q7.} For what tasks do you use AI in your current role?

\begin{itemize}
    \item[$\square$] Voice to Text
    \item[$\square$] Ambient Scribe
    \item[$\square$] Message drafting
    \item[$\square$] Patient communication/education
    \item[$\square$] Charting
    \item[$\square$] Literature search
    \item[$\square$] Decision Support
    \item[$\square$] Medication Management Support
    \item[$\square$] I do not use AI in my current role
    \item[$\square$] I am not sure
    \item[$\square$] Other (please specify):
\end{itemize}

\noindent\textbf{Definition: AI Agent}

An artificial intelligence (AI) agent is a system that autonomously performs tasks by designing workflows with available tools.\\

\noindent\textbf{Q8.} For non-medically complex patients, would you support or oppose AI agents operating as prescribers without requiring clinician sign-off?

\begin{table}[H]
\centering
\begin{tabular}{l *{3}{c}}
\toprule
 & \hspace{-0.5in}Oppose & \hspace{-0.1in}Neutral & \hspace{-0.1in}Support \\
\midrule
Refills for low-risk Rx & & & \\
Refills for high-risk Rx & & & \\
New prescriptions for low-risk Rx & & & \\
New prescriptions for high-risk Rx & & & \\
\bottomrule
\end{tabular}
\end{table}







\noindent\textbf{Q9.} For medically complex patients, would you support or oppose AI agents operating as prescribers without requiring clinician sign-off?

\begin{table}[H]
\centering
\begin{tabular}{l *{3}{c}}
\toprule
 & \hspace{-0.5in}Oppose & \hspace{-0.1in}Neutral & \hspace{-0.1in}Support \\
\midrule
Refills for low-risk Rx & & & \\
Refills for high-risk Rx & & & \\
New prescriptions for low-risk Rx & & & \\
New prescriptions for high-risk Rx & & & \\
\bottomrule
\end{tabular}
\end{table}








\subsection*{Calibrated Confidence-Based Escalation}

\textbf{Definition:} A calibrated confidence-based escalation mechanism is a mechanism where the agent automatically halts and routes a case to clinician review when its confidence falls below a defined threshold.\\

\noindent\textbf{Q10.} To what extent does the presence of a calibrated confidence-based escalation mechanism affect your willingness to permit AI agents to prescribe drugs?

\begin{itemize}
    \item[$\circ$] Not at all --- I would permit AI agents to prescribe drugs without it
    \item[$\circ$] Very little
    \item[$\circ$] Somewhat
    \item[$\circ$] To a great extent --- I would not permit AI agents to prescribe drugs without it
\end{itemize}

\noindent\textbf{Q11.} How do you rate the patient safety of each agent relative to an agent with no escalation mechanism?

\begin{table}[H]
\centering
\footnotesize
\begin{tabular}{l *{5}{c}}
  \toprule
  & \hspace{-0.6in}\shortstack{Much\\worse} & \hspace{-0.25in}\shortstack{Somewhat\\worse} & \hspace{-0.1in}\shortstack{About the\\same} & \hspace{-0.1in}\shortstack{Somewhat\\better} & 
  \hspace{-0.1in}\shortstack{Much\\better} \\
\midrule
Agent A (Calibrated) & & & & & \\

Agent B (Uncalibrated) & & & & & \\
\bottomrule
\end{tabular}
\end{table}






\noindent\textbf{Q12.} Who should bear primary responsibility for setting the confidence threshold at which an AI agent escalates to clinician review?

\begin{itemize}
    \item[$\circ$] AI company
    \item[$\circ$] Regulatory body (\eg FDA)
    \item[$\circ$] Hospital or health system
    \item[$\circ$] Independent clinical board
    \item[$\circ$] The clinician who will be reviewing
    \item[$\circ$] Multistakeholder body including patients/patient advocates
    \item[$\circ$] Combination or Other (please specify):
\end{itemize}

\subsection*{Communication of Confidence}

An AI agent is in place to prescribe new drugs and escalates a case for your review due to low confidence in its decision.\\

\noindent\textbf{Scenario A:} The AI agent is not confident because multiple drug options are clinically defensible and the evidence does not clearly favor one.\\

\noindent\textbf{Scenario B:} The AI agent is not confident because the patient's profile is rarely represented in its training data.\\

\noindent\textbf{Q13.} Do these two scenarios represent meaningfully different situations that should change how you approach the review?

\begin{itemize}
    \item[$\circ$] Yes, they require a fundamentally different approach
    \item[$\circ$] Yes, but only minor differences
    \item[$\circ$] No, I would approach both the same way
    \item[$\circ$] Unsure
    \item[$\circ$] Other (please specify):
\end{itemize}

\noindent\textbf{Q14.} For Scenario A, which of the following should the escalated interface show you?

\begin{itemize}
    \item[$\circ$] Final decision with no confidence information
    \item[$\circ$] Final decision clearly marked as low confidence
    \item[$\circ$] Summary of competing drug options with relative confidence values
    \item[$\circ$] Patient clinical data only
    \item[$\circ$] Unsure
    \item[$\circ$] Other (please specify):
\end{itemize}

\noindent\textbf{Q15.} For Scenario B, which of the following should the escalated interface show you?

\begin{itemize}
    \item[$\circ$] Final decision with no confidence information
    \item[$\circ$] Final decision clearly marked as low confidence
    \item[$\circ$] Summary of competing drug options with relative confidence values
    \item[$\circ$] Patient clinical data only
    \item[$\circ$] Unsure
    \item[$\circ$] Other (please specify):
\end{itemize}

\subsection*{Liability \& Accountability}

\textbf{Q16.} An AI agent issued a new prescription without communicating confidence or triggering escalation. The patient experienced a serious adverse drug reaction.

\noindent Rate responsibility for each party:

\subsection*{Q16A --- Low Confidence}

\begin{table}[H]
\centering
\begin{tabular}{l *{6}{c}}
\toprule
 & {0} & {1} & {2} & {3} & {4} & {5} \\
\midrule
AI company                 & & & & & & \\
Regulatory body            & & & & & & \\
Hospital or health system  & & & & & & \\
Overseeing clinician       & & & & & & \\
Patient                    & & & & & & \\
\bottomrule
\end{tabular}
\end{table}








\subsection*{Q16B --- High Confidence}

\begin{table}[H]
\centering
\begin{tabular}{l *{6}{c}}
\toprule
 & {0} & {1} & {2} & {3} & {4} & {5} \\
\midrule
AI company                 & & & & & & \\
Regulatory body            & & & & & & \\
Hospital or health system  & & & & & & \\
Overseeing clinician       & & & & & & \\
Patient                    & & & & & & \\
\bottomrule
\end{tabular}
\end{table}









\noindent\textbf{Q17.} An AI agent communicated confidence and triggered escalation. The clinician concurred and documented the rationale. The patient experienced a serious adverse drug reaction.

\subsection*{Q17A --- Low Confidence}









\begin{table}[H]
\centering
\begin{tabular}{l *{6}{c}}
\toprule
 & {0} & {1} & {2} & {3} & {4} & {5} \\
\midrule
AI company                 & & & & & & \\
Regulatory body            & & & & & & \\
Hospital or health system  & & & & & & \\
Overseeing clinician       & & & & & & \\
Patient                    & & & & & & \\
\bottomrule
\end{tabular}
\end{table}

\subsection*{Q17B --- High Confidence}

\begin{table}[H]
\centering
\begin{tabular}{l *{6}{c}}
\toprule
 & {0} & {1} & {2} & {3} & {4} & {5} \\
\midrule
AI company                 & & & & & & \\
Regulatory body            & & & & & & \\
Hospital or health system  & & & & & & \\
Overseeing clinician       & & & & & & \\
Patient                    & & & & & & \\
\bottomrule
\end{tabular}
\end{table}








\subsection*{Open-Ended Priorities (Optional)}

\textbf{Q18.} Is there anything about the use of AI agents in prescribing that you think the AI research community is not paying enough attention to?

\subsection{Appendix B: Additional Figures and Table}
\label{app:B}

\revision{
\begin{table}[H]
\centering
\small
\setlength{\tabcolsep}{1pt}
\begin{tabular}{@{}llrr@{}}
\toprule
\textbf{Characteristic} & \textbf{Category} & \textbf{n} & \textbf{\%} \\
\midrule
\multirow{3}{*}{Role (Q1)}
  & Physician                              & 102 & 75.0 \\
  & Nurse Practitioner                     &  30 & 22.1 \\
  & Physician Assistant                    &   4 &  2.9 \\
\midrule
\multirow{13}{*}{Specialty (Q2)}
  & Internal Medicine                      &  36 & 26.5 \\
  & Surgery                                &  17 & 12.5 \\
  & Pediatrics                             &  15 & 11.0 \\
  & Family Medicine/General Practice       &  13 &  9.6 \\
  & Neurology                              &  10 &  7.4 \\
  & Other                                  &  10 &  7.4 \\
  & Anesthesiology                         &   9 &  6.6 \\
  & Psychiatry                             &   7 &  5.1 \\
  & Obstetrics and Gynecology              &   4 &  2.9 \\
  & Physical Medicine and Rehabilitation   &   4 &  2.9 \\
  & Critical Care                          &   4 &  2.9 \\
  & Emergency Medicine                     &   3 &  2.2 \\
  & Radiology                              &   3 &  2.2 \\
\midrule
\multirow{2}{*}{\makecell{\noindent Prescriptive \\authority (Q3)}}
  & Full independent                       & 105 & 77.2 \\
  & Collaborative/restricted/limited       &  30 & 22.1 \\
\midrule
\multirow{5}{*}{\makecell{Years prescribing \\(Q4)}}
  & 1--5 years                             &  41 & 30.1 \\
  & 6--10 years                            &  22 & 16.2 \\
  & 11--15 years                           &  24 & 17.6 \\
  & 16--20 years                           &  14 & 10.3 \\
  & $>$20 years                            &  35 & 25.7 \\
\midrule
\multirow{2}{*}{\makecell{Uses AI in \\current role (Q7)}}
  & Yes                                    & 117 & 86.0 \\
  & No                                     &  19 & 14.0 \\
\bottomrule
\end{tabular}
\caption{Demographic composition of the analytic sample ($N=136$ U.S. prescribing clinicians), consolidating the per-item distributions for clinical role (Q1), specialty (Q2), prescriptive authority (Q3), years of prescribing experience (Q4), and current AI use (Q7). Percentages are of the full analytic sample; a small number of items had a single missing response, so column totals may not sum to exactly 100\%.}
\label{tab:demographics}
\end{table}
}

\begin{figure}[h]
    \centering
    \includegraphics[width=\columnwidth]{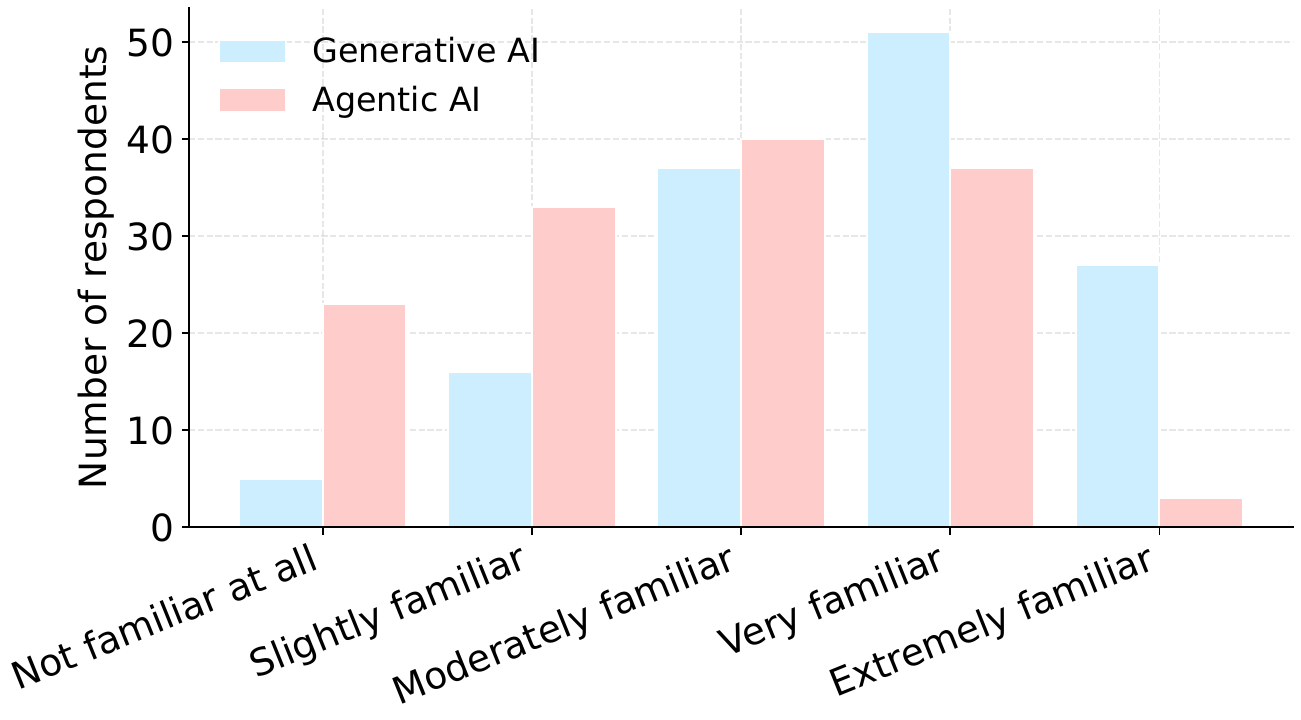}
    \caption{Distributed self-reported familiarity with generative AI vs. agentic AI. Results show that familiarity skews higher for generative AI but lower and more dispersed for agentic AI.}
    \label{fig:1}
\end{figure}

\begin{figure}[h]
    \centering
    \includegraphics[width=0.75\columnwidth]{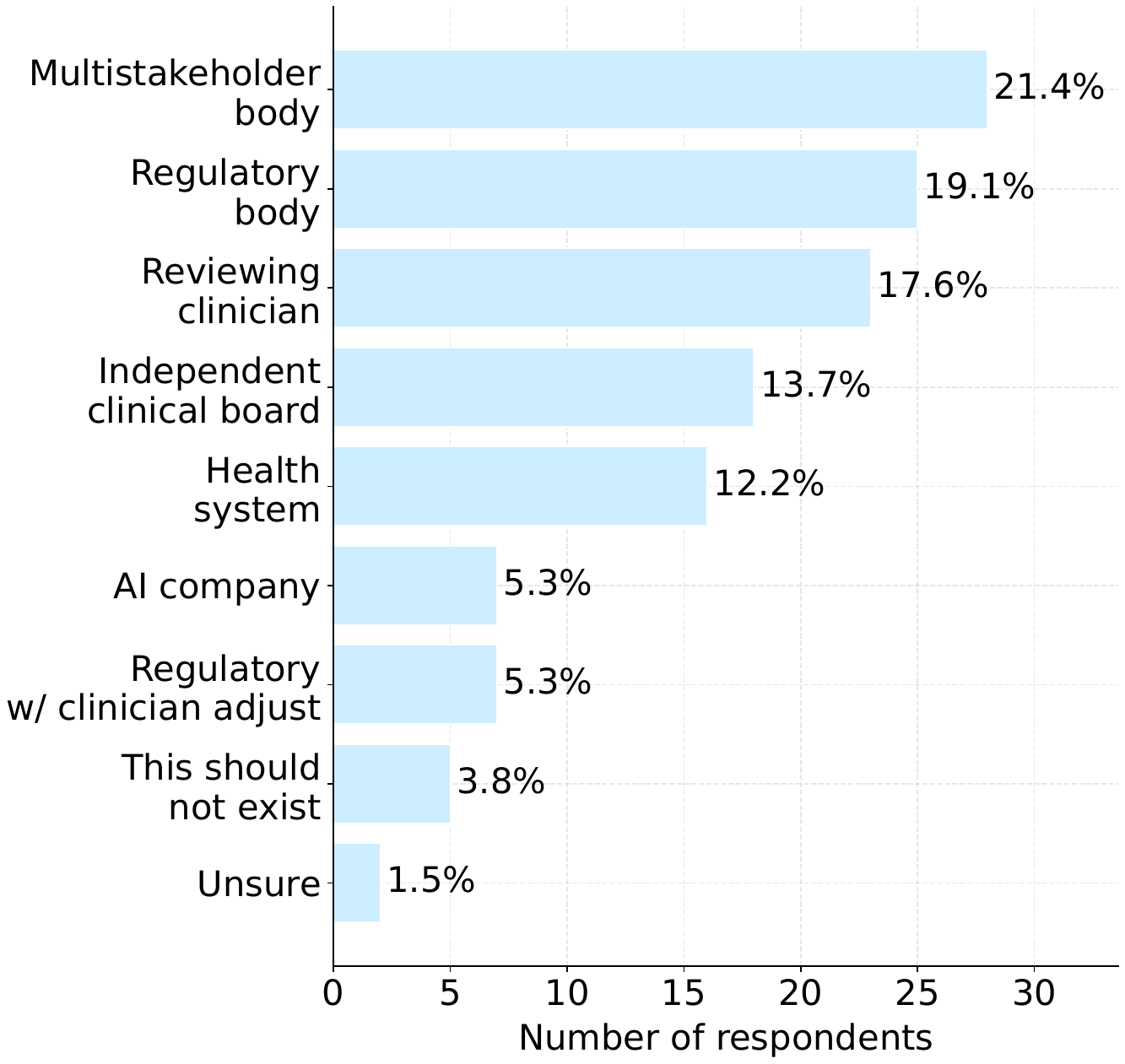}
    \caption{Preferred party for setting the confidence threshold at which an AI agent escalates to clinician review. Results show that no single party commanded a majority, but organization and oversight-level options combined account for 59.5\% of responses, while only 5.3\% of clinicians endorsed the AI company to set the threshold.}
    \label{fig:4}
\end{figure}

\subsection{Appendix C: Statistical Approach, Data Quality, and Additional Analyses}
\label{app:C}

\textbf{Demographics Exclusion.} We collected no personal demographics, including gender. As a fully anonymous survey of an IRB-designated vulnerable population (institutional employees), minimizing identifiers reduced re-identification risk in a small sample and reflected our view that such attributes were irrelevant to our hypotheses.

\noindent\textbf{Tests and analytic samples.} Formal hypothesis tests with reported $p$-values are restricted to the three predetermined hypotheses; all demographic breakouts and secondary contrasts are treated as descriptive and labeled exploratory. H1(a) was evaluated with a one-sided binomial test of the proportion endorsing calibrated escalation against a null of $0.5$ ($n=135$). H1(b) used a one-sided Wilcoxon signed-rank test comparing the calibrated agent to the uncalibrated agent ($n=133$). H2 used paired (two-sided) McNemar tests on the aleatoric versus epistemic interface choices ($n=130$). H3 used one-sided Wilcoxon signed-rank tests contrasting mean responsibility assigned to organizational versus individual parties within each scenario ($n=135$ for the no-escalation scenarios and $n=129$ for the escalation scenarios), with Friedman tests across the five parties as supporting evidence. Analytic $n$ varies slightly across items because it reflects the respondents who answered the specific item(s) for each test. We did not adjust for confounding variables as every predetermined test is a within-subjects (paired) comparison in which each respondent serves as their own control (With the exception of H1(a)).

\noindent\textbf{Completion and missingness.} Respondents were retained if they consented, pressed submit, held the ability to prescribe medications, and answered at least $80\%$ of survey questions. Only $4$ respondents answered fewer than $80\%$ of the questions and were removed from all analyses. Among retained respondents, no survey item exhibited a notable amount of missing data: the highest item-level missingness was $5.15\%$ for Q17 (the final adverse-outcome liability item), which is on the lower end of the normal range for a survey question, particularly given the comparatively high cognitive burden of that item and its position as the last substantive question of the survey. For the Q16/Q17 liability matrices, rows left entirely blank for a scenario were treated as missing, whereas rows partially completed were interpreted as ``not responsible'' ($0$) for the unrated parties, consistent with respondents rating only the parties they judged culpable.

\noindent\textbf{Multiplicity.} Where a single hypothesis is assessed with more than one confirmatory test, those tests are treated as one family and corrected together using the Holm--Bonferroni procedure, and we report both raw and Holm-adjusted $p$-values. This applies to H2, whose two McNemar contrasts (competing-options and abstention) form a family of size two, and to H3, whose four scenario-level directional contrasts (the low/high-confidence $\times$ escalation/no-escalation adverse-outcome scenarios) form a family of size four. For H1 we anchor a single primary confirmatory test per claim rather than correcting: H1(a) is anchored on the strongest-endorsement proportion (with the top-two-box proportion reported only as a supporting descriptive cut of the same item), and H1(b) is a distinct predetermined claim tested on a separate item, so neither enters a multiple-comparison family. The Friedman tests and the exploratory low- versus high-confidence within-party contrasts are descriptive, left uncorrected, and discussed as such. After Holm correction, all H2 and H3 confirmatory contrasts that were significant at the $0.05$ level remained so.

\noindent\textbf{Additional Analyses.} Beyond the three predetermined hypotheses, we conducted exploratory breakouts to test whether the central quantitative patterns varied with respondent background, specifically years of prescribing experience (Q4), clinical field, and self-reported familiarity with AI (Q5–Q6). We report these analyses descriptively, following our stated posture of reserving formal inference for H1–H3, with note that subgroups with fewer than ten respondents are flagged and not interpreted. Across all three demographic splits, we did not find evidence of meaningful subgroup differences on the core outcomes. In short, the willingness to require calibrated escalation, the safety advantage attributed to calibration, the differentiation of the escalated interface by uncertainty type, and the organization-over-individual liability pattern all appear to be broadly shared features of our clinician sample rather than artifacts of a particular experience level, specialty, or degree of AI exposure. We regard this stability as strengthening, rather than qualifying, the main findings, while noting that the small per-cell sample sizes mean these breakouts are best read as the absence of \textit{obvious} moderation rather than as positive evidence of none.

\subsection{Appendix D: Counterarguments}
\label{app:D}

\noindent\looseness=-1 A reasonable critique of our argument is that the architectural requirements we propose exceed what is currently demanded in comparable high-stakes domains. In medicine, therapies are routinely deployed on the basis of outcome evidence despite incomplete mechanistic understanding, and in other autonomous systems such as driving, regulation has emphasized constrained deployment and post hoc audit (\eg event recording and incident reporting) rather than per-decision interpretability. A related concern is that calibrated confidence and fine-grained uncertainty distinctions may be technically unstable in real-world clinical settings, where patient presentations often mix multiple sources of uncertainty, and where safety might instead be ensured through narrow scope restrictions and conservative operational design. While these critiques are sensible, our claim is not that interpretability or uncertainty decomposition must be perfect, but that once prescribing authority is delegated to an autonomous system, some form of per-decision, auditable uncertainty signal becomes load-bearing for safety and accountability. In this setting, calibrated escalation, uncertainty-aware interfaces, and prediction-level transparency function less as idealized technical desiderata than as minimal mechanisms for determining when the system should act, what it should communicate upon deferral, and how responsibility can be evaluated after the fact. Consistent with this framing, our survey results indicate that clinicians themselves treat these features not as optional enhancements but as preconditions for acceptable deployment.

\subsection{Appendix E: Data and Code}
\label{app:E}
\noindent In accordance with our IRB protocol and the anonymity assurances provided to participants, the underlying survey response data cannot be made publicly available, however the full analysis code and output is available at \url{https://anonymous.4open.science/r/agentic_AI_prescribing-41E2}.

\end{document}